\documentclass[letterpaper, 10 pt, conference]{ieeeconf}

\let\labelindent\relax
\usepackage{enumitem}

\usepackage{algorithm}
\usepackage{algpseudocode}
\usepackage{amsmath}
\usepackage{amssymb}
\usepackage{cite}

\usepackage{epsfig}
\usepackage{graphics}
\usepackage{graphicx}
\usepackage{mathptmx}
\usepackage{subcaption}
\usepackage{placeins}
\usepackage{blindtext}
\usepackage{times}

\newcommand\figref{Figure~\ref}

\IEEEoverridecommandlockouts
\title{\LARGE \bf
    Exposure Conscious Path Planning for Equal Exposure Corridors
}

\author{Eugene T. Hamzezadeh$^1$, John G. Rogers$^2$, Neil T. Dantam$^1$, and Andrew J. Petruska$^1$%
    \thanks{*This work was supported in part by the ARL TBAM-CRP [W911NF-22-2-0235]. \textit{(Corresponding author: Eugene Hamzezadeh.)}}%
    \thanks{$^1$ETH, NTD, and AJP are with Colorado School of Mines, Golden, CO 80401 USA \textit{(e-mail: \{ehamzeza, ndantam, apetruska\}@mines.edu)}}%
    \thanks{$^2$JGR is with the DEVCOM Army Research Laboratory, Adelphi, MD, 20783 USA \textit{(e-mail: john.g.rogers59.civ@army.mil)}}%
}

\begin{document}

\onecolumn
\hspace{0em} \vfill
\noindent \textbf{© 2024 IEEE. Personal use of this material is permitted. Permission from IEEE must be obtained for all other uses, in any current or future media, including reprinting/republishing this material for advertising or promotional purposes, creating new collective works, for resale or redistribution to servers or lists, or reuse of any copyrighted component of this work in other works.}

\vspace{1em}
\noindent Accepted to 2024 IEEE 20th International Conference on Automation Science and Engineering
\vfill \hspace{0em}
\newpage
\twocolumn

\maketitle
\thispagestyle{empty}
\pagestyle{empty}

\begin{abstract}
While maximizing line-of-sight coverage of specific regions or agents in the environment is a well-explored path planning objective, the converse problem of minimizing exposure to the entire environment during navigation is especially interesting in the context of minimizing detection risk. This work demonstrates that minimizing line-of-sight exposure to the environment is non-Markovian, which cannot be efficiently solved optimally with traditional path planning. The optimality gap of the graph-search algorithm A* and the trade-offs in optimality vs. computation time of several approximating heuristics is explored. Finally, the concept of \textit{equal-exposure corridors}, which afford polynomial time determination of all paths that do not increase exposure, is presented.
\end{abstract}

\section{Introduction}

\begin{figure}[t]
    \centering
    \includegraphics[width=\linewidth]{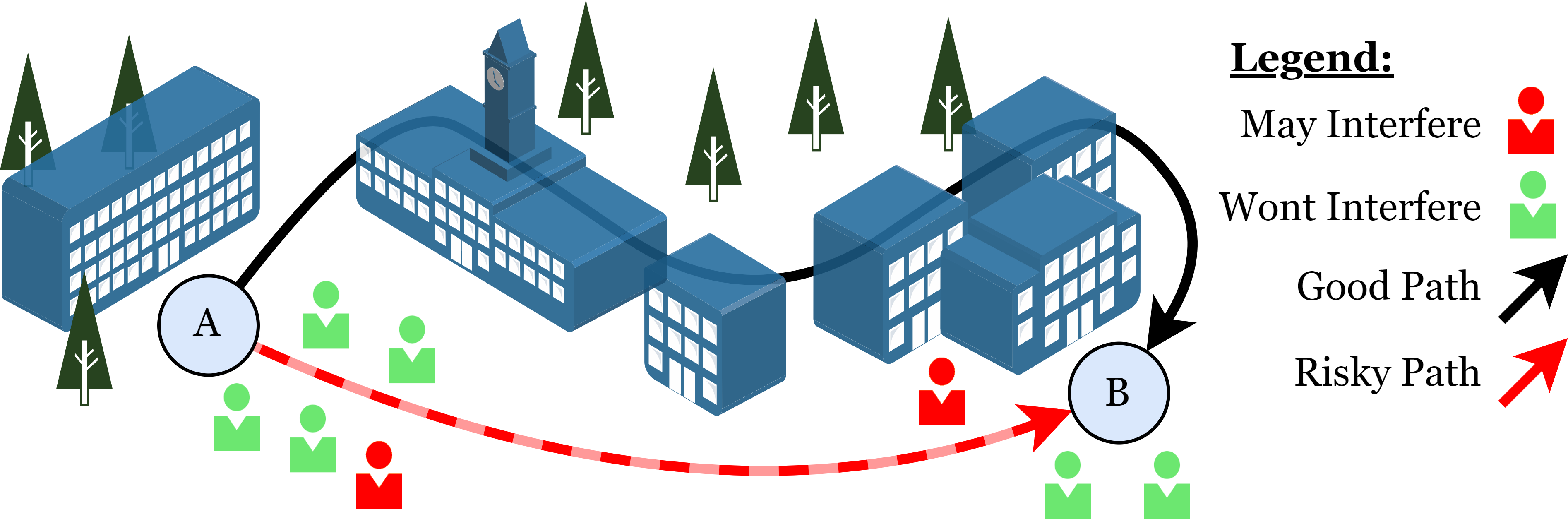}
    \caption{When delivering a package the robot should take the solid black line route over the dashed red one, so as to minimize the likelihood of being seen by a malicious agent. \vspace{-1em}}
    \label{fig:motivating_example}
\end{figure}

Robots are subject to several exposure mechanisms that path planners consider to improve system performance. Traditional exposure mechanisms such as radiation \cite{lee_2024_static_path_plan_minimize_radiation, liu_2014_particle_swarm_radiation_path_planning} and signal strength or interference \cite{remley2007nist_search_rescue, schack2021information_gain} are often driving metrics for the quality and feasibility of paths. Path planners may instead consider indirect exposure such as visual line-of-sight. While many scenarios such as robot exploration or search and rescue maximize line-of-sight, this work focuses on minimizing line-of-sight, which is relevant to the service robot navigation scenario presented in \figref{fig:motivating_example}.

Robots show potential for automating mundane service tasks, yet struggle with human interference in public environments \cite{salvini_2010_how_safe_are_service_robots_bullying_robots, keijsers_2018_mindless_robots_get_bullied}. In the delivery scenario, the presented algorithms reduce visual exposure to the environment, minimizing delivery disturbances and the probability of discovery by malicious actors. Besides automated last-mile delivery, our results are relevant to robotic maintenance tasks and remote wildlife observation scenarios among others.

Prior work has focused on robot exploration that maximizes line-of-sight coverage for information gain while minimizing the path or time taken to explore the entire environment \cite{lubanco_2020_review_of_utility_cost_exploration}. While many solutions address the information gain problem that optimizes path length  \cite{schack2021information_gain, martin_2008_non_markovian_ground_search}, these do not directly apply to the converse problem of minimizing exposure without any regard for path length. Minimizing exposure along a path fundamentally depends on the path taken in full, and not solely on the transitions along the path, which makes the problem non-Markovian when the state is only the present robot position. Encoding information about the history of configurations would make the problem Markovian, but requires a state-space the dimension of all possible robot positions, and the curse of dimensionality would quickly make the problem intractable \cite{bellman1959adaptive}.

Common path planning approaches include search-based methods, sampling-based methods, and optimization techniques \cite{lynch_park_2017_modern_robotics_planning_and_control}. Classical heuristic search-based planners such as Dijkstra's algorithm \cite{dijkstra1959}, A* \cite{astar_hart1968}, AD* \cite{aine_2013_anytime_d_star}, and ARA* \cite{likhachev_2203_ara_star} provide optimal paths, but constructing effective heuristics to scale to high-dimensional problems can present challenges. Sampling-based methods scale well to high-dimensional problems but only offer probabilistic completeness \cite{lavalle_1998_rapidly_exploring_random_trees} or asymptotic optimality \cite{karaman_2011_sampling_based_algs_optimal_motion_planning}. Optimization-based approaches such as CHOMP \cite{ratliff_CHOMP}, STOMP \cite{kalakrishnan_STOMP}, and TrajOpt \cite{schulman_TrajOpt} only guarantee optimality locally and require an initial path.

Movement corridors, which define the permissible movement area, have been explored in the context of path planning to enable local planners to efficiently account for dynamic obstacles without having to collision check against static ones. While the corridor size is typically represented as a function of the distance to the nearest obstacle \cite{wang_2015_corridor_based_local_planning} \cite{geraerts_2007_corridor_map_method}, other movement corridor methods, such as in \cite{schitz_2021_corridors_teleoperated_driving}, maximize natural planning intuition with human-defined boundaries. The current methods, however, do not extend this concept to equal exposure or equal objective cost corridors.

The first important contribution of this work is demonstrating exposure minimization is non-Markovian and quantifying the optimality gap of several Markovian-approximating graph-search implementations, for which A* is chosen as it remains one of the most applied algorithms for path planning \cite{sedighi2019_hybrid_astar_parking, kuswadi_2018_a_star_slam_application, kusuma_2019_a_star_humanoid_robots_application_example} and is frequently a fundamental aspect of more contemporary approaches \cite{li2020_genetic_algorithm_astar_heuristic, lan2021_robot_pathplanning_antcolony_astar_algorithm,  yong_2023_improved_hector_slam_navigation_strategy, chen_2023_improved_a_star_method}. The second important contribution is presenting \textit{equal-exposure corridors}, which define the set of all paths that are equal in exposure cost to the exposure-conscious path. Local planners can leverage these corridors to navigate in dynamic environments without negatively contributing to the exposure metric. Team-based planners can use corridors to quickly compute paths for subsequent robots within the same exposure constraints.

The rest of this paper is organized as follows. Section \ref{sec:Methods} formulates the problem, describes the search metrics, and details corridor construction. Section \ref{sec:Experimental} discusses the experimental setup, with results provided in Section \ref{sec:Results}, and commentary with future directions presented in Section \ref{sec:Discussion}.

\section{Methods} \label{sec:Methods}

\subsection{Problem Formulation}
\begin{figure}[t]
    \vspace{0.5em}
    \centering
    \begin{subfigure}[t]{0.17944751381\textwidth}
        \centering
        \includegraphics[width=\textwidth]{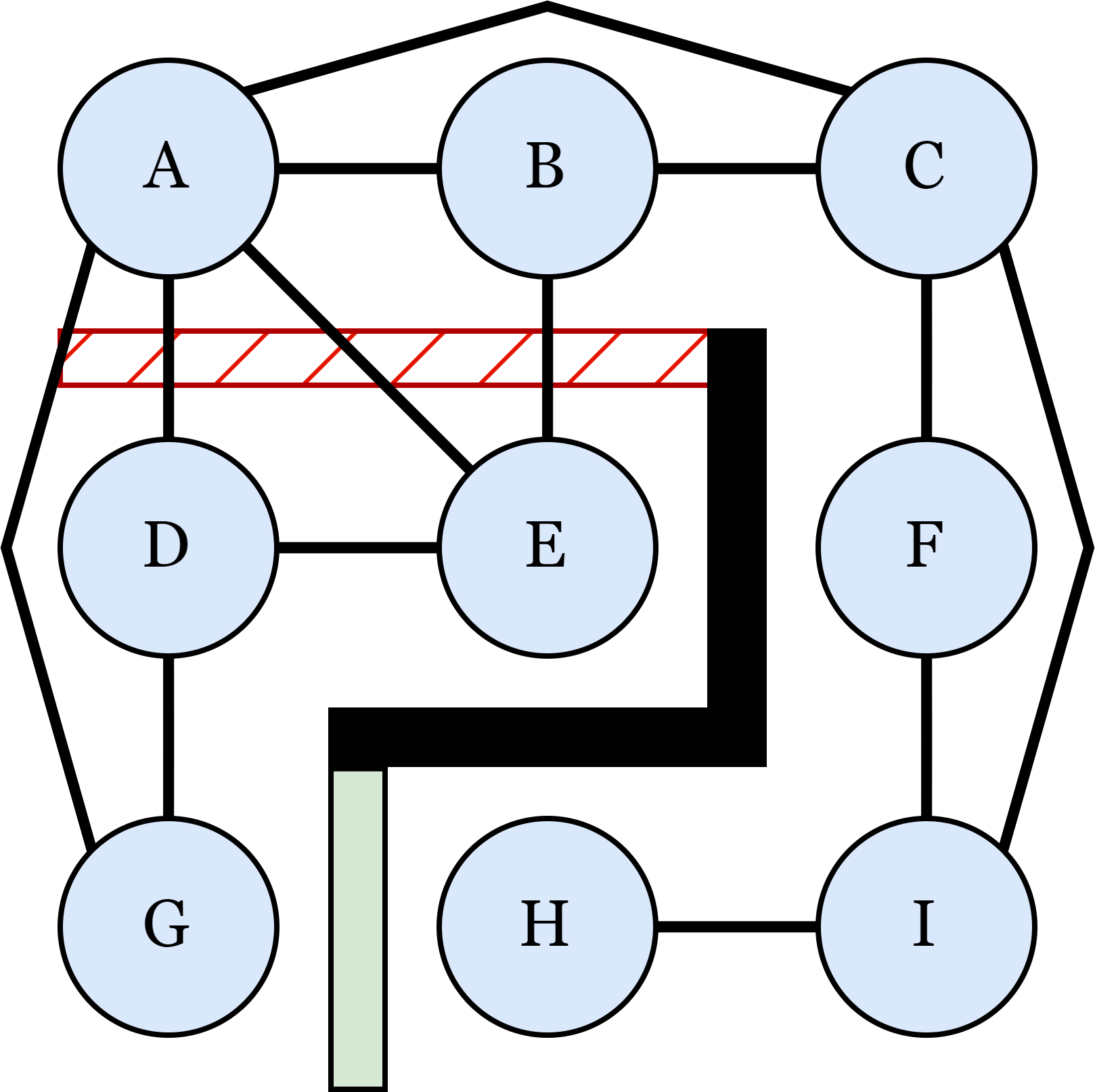}
        \caption{Exposure Graph}
        \label{fig:exposure_graph_example}
    \end{subfigure}%
    \hspace{0pt}
    \begin{subfigure}[t]{.16\textwidth}
        \centering
        \includegraphics[width=\textwidth]{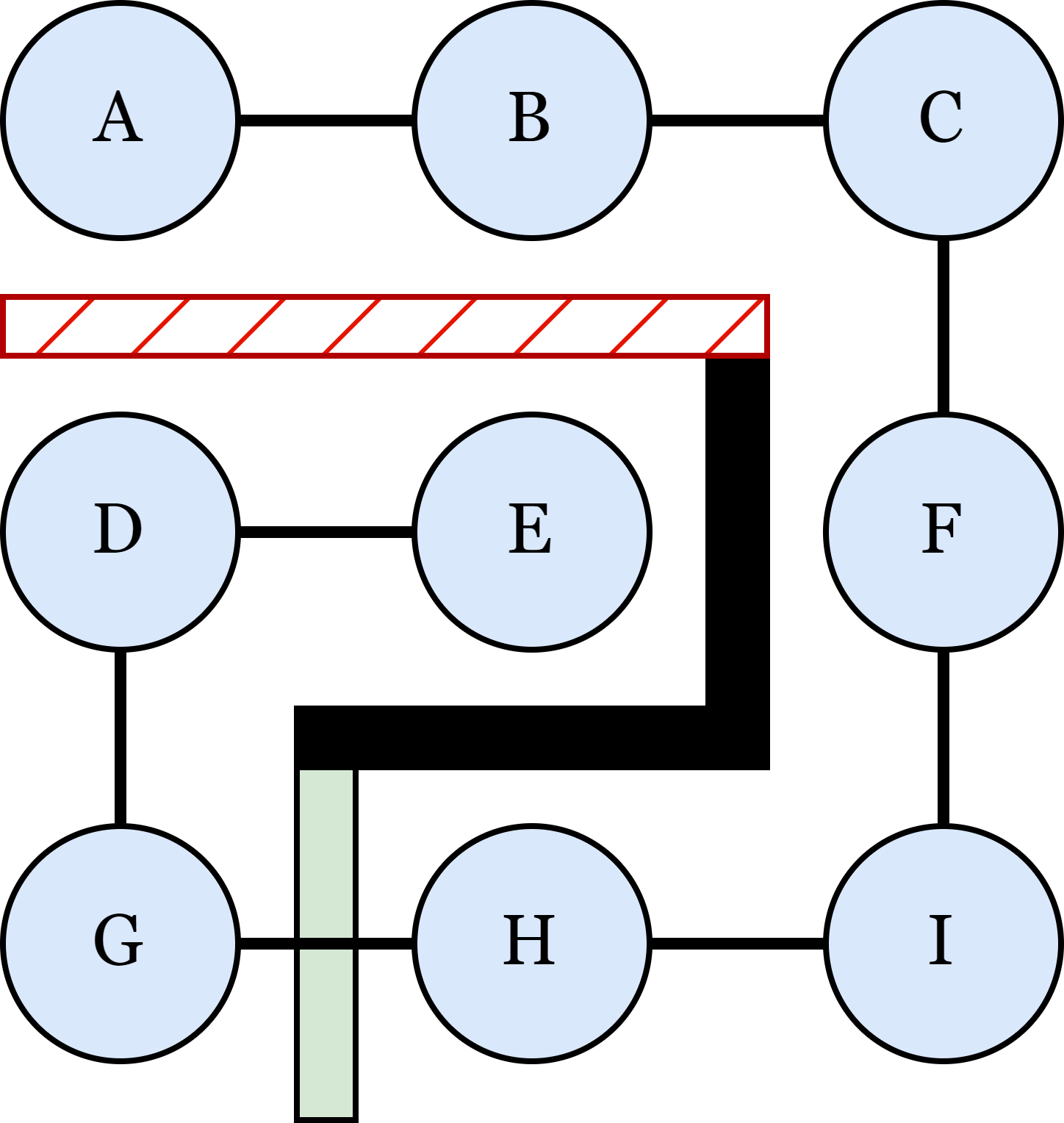}
        \caption{Traversability Graph}
        \label{fig:traversability_graph_example}
    \end{subfigure}
    \hspace{0pt}
    \begin{subfigure}[t]{.125\textwidth}
        \centering
        \includegraphics[width=\textwidth]{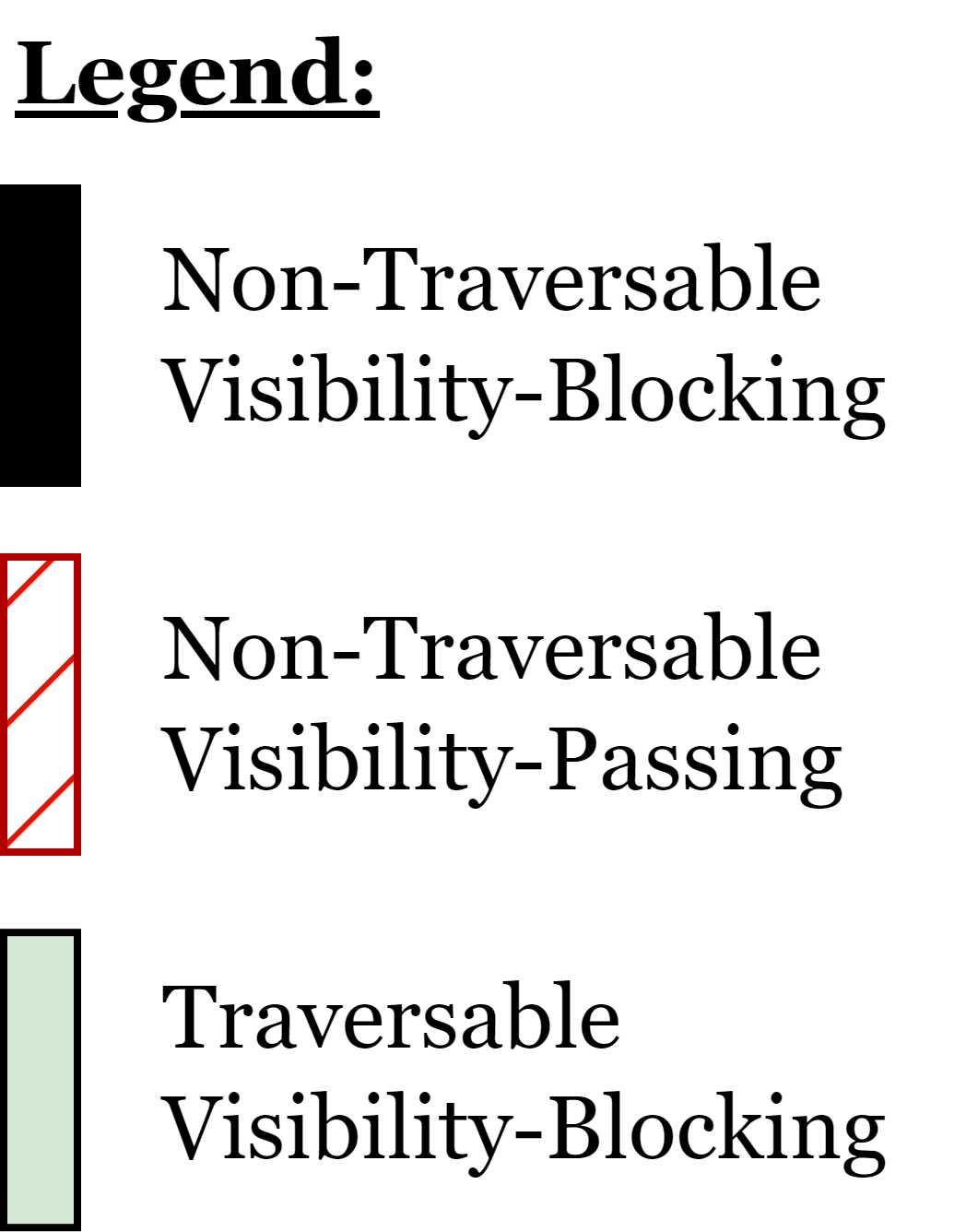}
    \end{subfigure}
    \hspace{0pt}

    \caption{Example graph formulation demonstrating an exposure graph (a) and traversability constraint graph (b) for a problem where the exposure function is visual line-of-sight. \vspace{-1em}}
    \label{fig:example_traversability_visibility_graphs}
\end{figure}

Objective: Find path $P$, of any length, that minimizes the visual exposure of a single agent to the environment $E$ during the traversal between a starting point in region $R_s$ and goal point in region $R_g$ while satisfying additional constraints, i.e.\,traversability. This formulation constitutes a set of two disjoint graphs, one encoding the agent's constraint-free transitions and one encoding the exposure function between each region, as depicted in \figref{fig:example_traversability_visibility_graphs}, where the regions of the environment are the nodes and the values of the traversability and exposure functions are the edges of each graph, respectively.

\begin{description}[labelsep=1em, leftmargin=0.125in]
    \item [Definition 1.1] (Environment):
    The environment $E$ is a collection of $n$ discrete regions $E = \{R_1, R_2, ..., R_n\}$ within a certain area of interest. Every region $R_i$ contains a single representative point $P_i = (x_i, y_i, z_i)$.
    \item [Definition 1.2] (Traversability Function):
    The traversability function $\text{\textbf{\textit{trav}}}(R_a, R_b) \rightarrow \mathbb{B}$ encodes the existence of a path between regions $R_a$ and $R_b$ that does not exceed the capabilities of the system and does not require the passage through any other region.
    \begin{equation}
    \text{\textbf{\textit{trav}}}(R_a, R_b) = 
        \begin{cases}
            1 & \text{if $\text{\textbf{\textit{adj}}}(R_a, R_b)$ and $\text{\textbf{\textit{cap}}}(P_a, P_b)$}\\
            0 & \text{otherwise}
        \end{cases}
    \end{equation}
    Where, $\text{\textbf{\textit{adj}}}(R_a, R_b)$ is true if regions $R_a$ and $R_b$ are physically adjacent and $\text{\textbf{\textit{cap}}}(P_a, P_b)$ is true if the robot is mechanically capable of bidirectionally moving between $P_a$ and $P_b$. The edges of the constraint graph are constructed for every combination of regions $(R_a, R_b): R_a, R_b \in E$ where the constraint function $\text{\textbf{\textit{trav}}}(R_a, R_b)$ is true.
    \item [Definition 1.3](Exposure Function):
    The exposure function $\text{\textbf{\textit{exp}}}(R_a, R_b) \rightarrow \mathbb{B}$ encodes the existence of visual line-of-sight between the specified regions $R_a$ and $R_b$.
    \begin{equation}
    \text{\textbf{\textit{exp}}}(R_a, R_b) = 
        \begin{cases}
            1 & \text{if } R_a \text{ and } R_b \text{ share line-of-sight}\\
            0 & \text{otherwise}
        \end{cases}
    \end{equation}
    The edges of the exposure graph are constructed for every combination of regions $(R_a, R_b): R_a, R_b \in E$ where the exposure function $\text{\textbf{\textit{exp}}}(R_a, R_b)$ is true.
    \item [Definition 1.4] (Exposure Set):
    We define the exposure set $\mathcal{E}(x)$ of a region $R_x$ as the set of all other regions in the environment within line-of-sight, that is
    \begin{equation}
        \label{region_exposure_set}
        \mathcal{E}(x) = \{R_k | R_k \in E, \text{\textbf{\textit{exp}}}(R_k, R_x) = \mathrm{true}\}.
    \end{equation}
\end{description}

Without loss of generality, we make the following mild assumptions about region properties and how a region's representative point defines its line-of-sight.

\begin{description}[labelsep=1em, leftmargin=0.125in]
    \item [Assumption 1:] Every region is exposed to itself and encompasses the free configuration space for that area.
    \item [Assumption 2:] The representative point $P_x$ of a region $R_x$ contains enough information to fully describe the line-of-sight region $R_x$ experiences.
    \item [Assumption 3:] Every point $Q$ contained in the region $R_x$ has the exact same line-of-sight information as the region's representative point $P_x$.
\end{description}

\subsection{Objective Functions}

Two forms of line-of-sight exposure are considered--binary and accumulative. Binary exposure only considers if a region is ever exposed to the path, while accumulative exposure considers repeated exposure to a region as additional exposure. In the binary sense, the robot only risks failure the first time it exposes a new region. In the accumulative sense, the robot repeatedly risks failure for every exposure. With either exposure concept, the objective is to find a path $P = \{P_1, P_2, ..., P_L\}$ that satisfies the traversability requirement and minimizes the desired exposure cost function. It is important to note that path length is not considered, as a tortuous route through a ravine could avoid significant exposure while a direct route across a plane could incur significant exposure.

\subsubsection{Traversability Constraint}
To satisfy the traversability constraint, the path $P = \{P_1, P_2, ..., P_L\}$ must start in region $R_s$, end in the goal region $R_g$, and pass the traversability function for every transition along the path, that is
\begin{equation} 
    \text{\textbf{\textit{trav}}}(P_i, P_{i+1}) = \mathrm{true} \text{ } \forall i \in [1, L-1].
\end{equation}

\subsubsection{Binary Objective Function} The objective function to minimize in a binary environment is defined as the total number of regions that share line-of-sight with any step of the path $P$. Binary algorithms minimize the function
\begin{equation}
    \label{binary_cost_function}
    \textbf{\textit{obj}}_{bin}(P) = | \bigcup_{P_i\in P}\ \mathcal{E}(i) |,
\end{equation}
where the operator $|\cdot|$ denotes set size.

\subsubsection{Accumulative Objective Function} The accumulative cost function requires the probability of continued success $p_{success}$ after each exposure. The objective function is
\begin{equation}
    \label{accumulative_cost_function}
    \textbf{\textit{obj}}_{acc}(\mathcal{A}(P_b)) = \sum_{c_i \in \mathcal{A}(P_b)} - \log_{10}(\max(p_{success}^{c_i}, p_{success}^\tau),
\end{equation}
where $c_i$ is the number of times region $R_i$ shared line-of-sight with the path, stored in the set $\mathcal{A}(P_b)$, and $\tau$ is the number of exposures at which region $R_i$ no longer affects the metric.

\subsection{Non-Markovian Space}
\noindent \textbf{Lemma 1:} Optimal binary global exposure paths need not consist of optimal sub-paths.
\vspace{4pt}

\begin{figure}[t]
    \vspace{5pt}
    \centering
    \includegraphics[width=\linewidth]{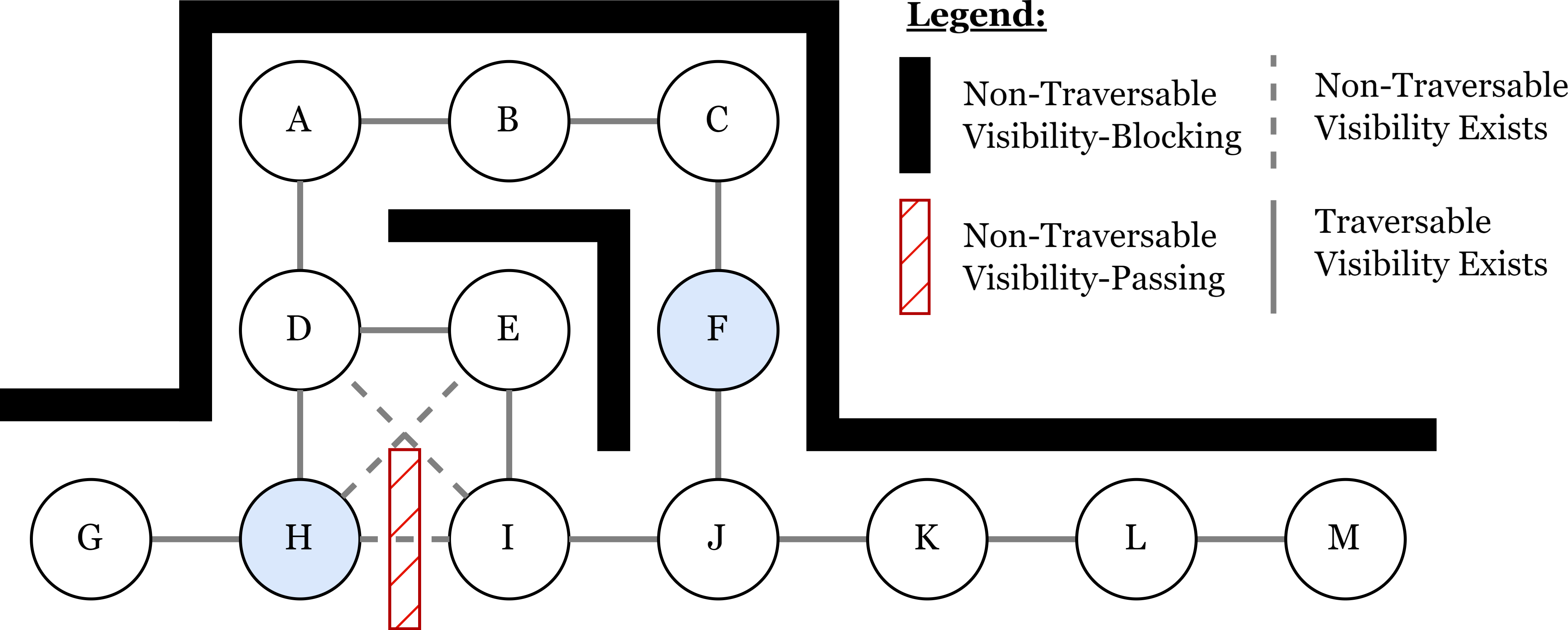}
    \caption{An example line-of-sight exposure scenario where the optimal path $F \rightarrow H$ does not contain the optimal sub-path $F \rightarrow E$ when the objective is to minimize the number of positions line-of-sight is shared with. \vspace{-1em}}
    \label{fig:example_not_suboptimal_paths}
\end{figure}

\noindent \textbf{Proof:}
Consider the environment presented in \figref{fig:example_not_suboptimal_paths}. The solid-black obstacles block visibility and movement, while the red-hatched obstacles block only movement. The objective is to minimize the number of positions visible when moving from location $F$ to $H$. This leads to the following:
\begin{enumerate}
    \item The exposure optimal path $P_1$ from $F$ to $H$ would take $\{F \rightarrow J \rightarrow I \rightarrow E \rightarrow D \rightarrow H\}$ exposing $12$ positions $(A, C, D, E, F, G, H, I, J, K, L, M)$.
    \item The sub-path of $P_1$ from $F$ to $E$, $P_{1, F \rightarrow E}$ consists of $ \{F \rightarrow J \rightarrow I \rightarrow E\}$ and exposes $11$ positions $(C, D, E, F, G, H, I, J, K, L, M)$.
    \item The sub-path of $P_1$ from $E$ to $H$, $P_{1, E \rightarrow H}$ consists of $\{E \rightarrow D \rightarrow H\}$ and exposes $10$ positions $(A, D, E, G, H, I, J, K, L, M)$.
    \item The exposure optimal path $P_2$ from $F$ to $E$ is $\{F \rightarrow C \rightarrow B \rightarrow A \rightarrow D \rightarrow E\}$ and exposes only $9$ positions $(A, B, C, D, E, F, H, I, J)$.
\end{enumerate}
Therefore, since the optimal path $P_2$ outperforms the sub-path $P_{1, F \rightarrow E}$ of the total path $P_1$, sub-paths of an optimal path are not in themselves guaranteed to be optimal.

Without optimal substructure, Bellman's condition is not satisfied, and standard graph-search algorithms (e.g. Dijkstra's, A*) cannot guarantee optimality \cite{salzman2017efficient}. Further, since the sum of the individual sub-path costs $P_{1, F \rightarrow E}$ and $P_{1, E \rightarrow H}$ does not equal the total path cost of $P_1$, exposure is therefore non-Markovian if position is taken as the state. However, Bellman's condition and the Markovian nature are recouped if, instead, the traversed path (all positions visited) is retained as the state, which greatly increases the problem dimension.

\subsection{Search Algorithms}

\subsubsection{Best First Search}
To compare to optimal path solutions, we implement a best-first-search algorithm \cite{russell_norvig_2016_artificial_intelligence_a_modern_approach} with a modified search state $S = \{R_s, V\}$. The modified search state consists of the search node's region $R_s$ and a set of previously visited regions $V$. 
This enables the algorithm to evaluate the cost of the path as a whole, producing optimal results at the expense of an exponential increase in search space. The cost function resembles equation \eqref{binary_cost_function} and is the number of regions exposed by visiting each element of $V$, i.e.
\begin{equation}
    \textbf{\textit{cost}}_{bfs}(S) = | \bigcup_{R_v \in V} \mathcal{E}(v) |.
\end{equation}

The heuristic function between a state $S$ and the goal $R_g$ is defined as the number of new regions exposed by the goal that have not been exposed by any of the previously visited regions in $V$, i.e.
\begin{equation}
    h_{bfs}(S, R_g) = | \{ R_i | R_i \in \mathcal{E}(g) \land R_i \not\in \epsilon_V \} |,
\end{equation}
where $\epsilon_V$ is the set of all regions exposed by any region $R_v$ in the visit set $V$, i.e.
\begin{equation}
    \epsilon_V = \bigcup_{R_v \in V} \mathcal{E}(v).
\end{equation}

The memory footprint and computation time of this best-first-search approach make it impractical for real-time application to mobile systems. Even offline, this algorithm scales poorly and becomes intractable for longer paths.

\subsubsection{A* Exposure Score Search}
This variant of A* seeks to find a balance between computation time and path quality with the use of a fixed pre-computation period to assign transition costs for each adjacent region transition. Every region $R_i$ in the environment $E$ is given an exposure score $e_i \in [0, 1]$ which is defined as 
the proportion of the environment that region $R_i$ has line-of-sight with
\begin{equation}
    e_i = \frac{| \mathcal{E}(i) |}{n}.
    \label{eqn:exposure_score}
\end{equation}
The transition cost function closely resembles a shortest-path A* search and is defined as
\begin{equation}
    t_{ess}(R_a, R_b) = e_b,
\end{equation}
where $R_a$ and $R_b$ are adjacent regions and $e_b$ is the exposure score of the destination region. The heuristic cost function between the current region $R_b$ and goal $R_g$ is
\begin{equation}
    h_{ess}(R_b, R_g) = M(P_b, P_g) \cdot \delta,
\end{equation}
where $\delta$ is the minimum exposure score for any region in the environment 
\begin{equation}
\delta = \min(\{e_i|, i \in [1, n]\}),
\end{equation}
and $M(P_b, P_g)$ is the Manhattan distance between the representative points of each region 
\begin{equation}
    M(P_b, P_g) = |P_{b,x} - P_{g,x}| + |P_{b,y} - P_{g,y}| + |P_{b,z} - P_{g,z}|.
\end{equation}

\subsubsection{A* Binary Search}
The A* Binary variant minimizes the total number of positions that are exposed by the path in a Boolean sense. Once a position is exposed, additional exposures do not contribute to the path cost. We define the search node $N_{bin} = (R, \mathcal{B}(P), N_{bin,parent})$ where $R$ is the region of the search node, $\mathcal{B}(P)$ is the binary exposure accumulator, which stores all the regions exposed by that the path taken so far. The root node of the search tree is
\begin{equation}
    N_{s} = (R_s, \mathcal{E}(s), \varnothing),
\end{equation}
where the root node's exposure accumulator $\mathcal{B}(P)$ is the same as the exposure set $\mathcal{E}(s)$ for the start region $R_s$. When a node $N_p$ is expanded, every child node $N_c$ of adjacent region $R_c$ is constructed as follows
\begin{equation}
    N_c = (R_c, \mathcal{B}(P) \cup \mathcal{E}(c), N_p).
\end{equation}
Note: the exposure accumulator of a child node always contains the exposure set of \textbf{all} its parent nodes. That is, for any parent search node $N_p$ and child search node $N_c$, $\mathcal{B}(P_p) \subseteq \mathcal{B}(P_c)$
the transition cost function $t_{bin}(N_a, R_b)$ is similar to equation \eqref{binary_cost_function} and is
\begin{equation}
    t_{bin}(N_a, R_b) = | \mathcal{B}(P_a) \cup \mathcal{E}(b) | - | \mathcal{B}(P_a) | + m,
\end{equation}
where $N_a$ is the search node, $R_b$ is the adjacent region, and $m$ is a tiny movement cost to discourage the algorithm from oscillating in areas where $\mathcal{B}(P_a) \cup \mathcal{E}(b)$ produces no additional exposures. Intuitively, the transition cost to an adjacent region is the number of \textit{new} positions that the step introduces. The value of $m$ is chosen such that for it to outweigh one additional exposure, the algorithm would have to traverse every region in the environment.

The heuristic cost function $h_{bin}(N_b, R_g)$ between a search node $N_b$ and the goal region $R_g$ is defined as the number of positions the goal exposes that the path thus far has not exposed. An admissible lower bound is the best-case scenario for a path trying to reach the goal and is
\begin{equation}
    h_{bin}(N_b, R_g) = | \{ R_i | R_i \in E, R_i \in \mathcal{E}(g) \land R_i \not\in \mathcal{B}(P_b) \} |.
\end{equation}

\subsubsection{A* Saturation Search}
The success of a path does not always depend solely on binary exposure. Sometimes the number of exposures (or equivalently length of exposure) to a region is of significance. A* Saturation uses the probability of success after an exposure $p_{success}$ and an exposure saturation threshold $\tau$ to define how multiple exposures contribute to path outcome. The exposure saturation threshold $\tau$ defines the exposure count at which a region saturates, and further exposure to the region does not continue to affect the outcome. The appropriate selection of $\tau$ allows the search to consider exposure scenarios between A* Exposure Score and A* Binary. Notably, 
if $\tau = 1$, the search is equivalent to the binary search scenario with a different heuristic. Otherwise, as $\tau \rightarrow n$, where $n$ is the number of regions in the environment, the search resembles A* Exposure Score. Occupying a region fully saturates its exposure value to $\tau$.

The node structure $N_{sat}$ of A* Saturation is similar to A* Binary, but that accumulator is replaced with a structured set of non-negative integer values. Thus, a node is defined as $N_{sat} = (R, \mathcal{A}(P), N_{sat, parent})$, where $\mathcal{A}(P)$ is the accumulator that sums the number of times each region $R_i$ is exposed by the path $P$. The root node $N_{s}$ from the starting region $R_s$ is
\begin{equation}
    N_{s} = (R_s, \{ 1 \text{ if } R_i \in \mathcal{E}(s) \text{ else } 0  \text{ } \forall i\}, \varnothing).
\end{equation}

When a node $N_p$ is expanded, every child node $N_c$ of adjacent region $R_c$ is constructed as follows
\begin{equation}
    \label{sat_node_equation}
    N_c = (R_c, \mathcal{A}(P) + \left\{ \begin{array}{l}
        1 \text{ if } R_i \in \mathcal{E}(c) \text{,}\\
        \tau \text{ if } R_i = R_c \text{,}\\
        0 \text{ otherwise }
    \end{array}
    \text{~} \forall i\right\} , N_p).
\end{equation}
The transition cost $t_{sat}(N_a, R_b)$ between a search node $N_a$ and an adjacent region $R_b$ is 
\begin{equation}
    t_{sat}(N_a, R_b) = \textbf{\textit{obj}}_{acc}(\mathcal{A}(P_b)) - \textbf{\textit{obj}}_{acc}(\mathcal{A}(P_a)),
\end{equation}
where $\mathcal{A}(P_b)$ is constructed according to equation \eqref{sat_node_equation} and the function  $\textbf{\textit{obj}}_{acc}(\mathcal{A}(P))$ is defined in equation \eqref{binary_cost_function}. The heuristic from $R_b$ to the goal $R_g$ is defined as
\begin{equation}
    h_{sat}(R_b, R_g) = - M(P_b, P_g) \cdot \tau \log_{10}(p_{success}).
\end{equation}

\subsection{Equal-Exposure Corridors}
After a path $P$ has been computed from region $R_a$ to $R_b$ we construct a set $K$ which represents every region that was exposed along the path as follows. Note that the regions along a path are always contained within the total exposure set ($P \subseteq K$), i.e.\,
\begin{equation}
    \label{exposed_by_path}
    K = \bigcup_{P_i\in P}\{R_j | R_j \in E, \text{\textbf{\textit{exp}}}(R_j, P_i) = \mathrm{true}\}.
\end{equation}

The corridor of equal exposure $C$ is then defined as the set of all regions that do not share line-of-sight with any region not contained in $K$, i.e.\,
\begin{equation}
    C = \{R_i | R_i \in E, \text{\textbf{\textit{exp}}}(R_t, R_i) = \mathrm{false} \text{ } \forall R_t \in \bar{K} \}.
\end{equation}
That is, $C$ contains every location in the environment that could be visited while not increasing the path's total exposure. This computed corridor $C$ can then be used to express \textbf{\textit{every}} path which could be taken in the environment $E$ that would not negatively impact the global exposure.

\begin{figure}[b]
    \centering
    \begin{subfigure}[b]{.217\textwidth}
        \centering
        \includegraphics[width=\textwidth]{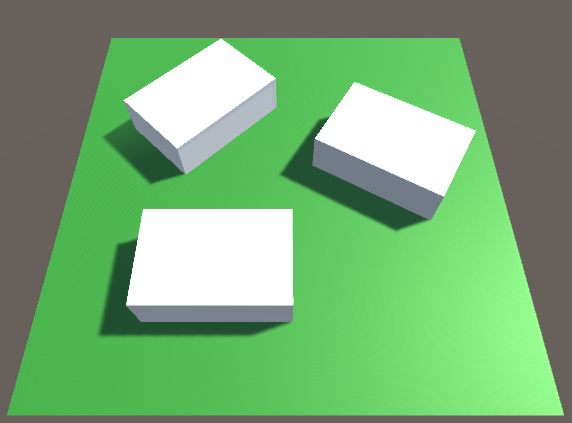}
        \caption{Boxes Environment}
        \label{fig:boxes_overview}
    \end{subfigure}%
    \hspace{0pt}
    \begin{subfigure}[b]{.25\textwidth}
        \centering
        \includegraphics[width=\textwidth]{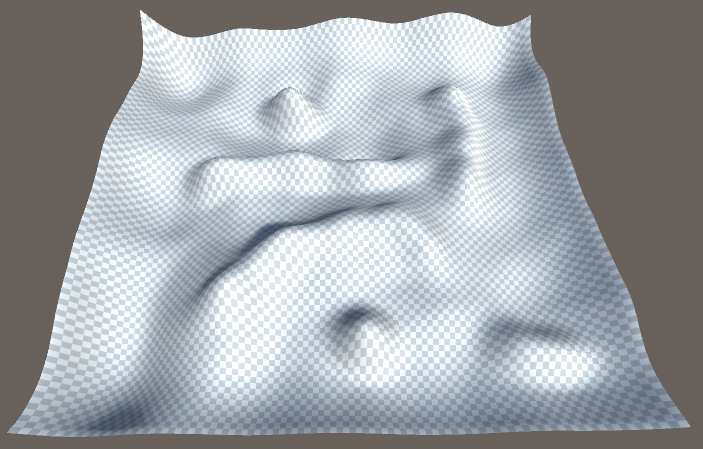}
        \caption{Hills Environment}
        \label{fig:Hills_overview}
    \end{subfigure}
    \caption{Overview of the environments tested. Boxes (a) provides large areas of zero-gradient and gradient spikes. Hills (b) supplies a well-defined gradient but several local exposure minima.}
    \label{fig:map_overviews}
\end{figure}

\section{Experimental Setup}\label{sec:Experimental}
The three A* based algorithms were tested in two environments, Boxes and Hills -- Figures \ref{fig:boxes_overview} and \ref{fig:Hills_overview} respectively, which were specifically chosen because they experience highly-contrasting exposure gradients as a function of local movement. The Boxes environment experiences large portions of zero-gradient due to the general flatness of the terrain combined with exposure-cost spikes. Approaching a box corner has little exposure change, but moving around a corner of one of the boxes drastically increases it. Conversely, the Hills environment has well-defined non-zero exposure gradients for the majority of the map but has several local exposure minima.

We construct the environment $E$ as a grid of equal-sized square regions and choose each region's representative point $P_i = (x_i, y_i, z_i)$ such that $(x_i, y_i)$ is directly in the middle of the grid cell and $z_i$ is a non-negative offset $d$ upwards from the highest point of the topological surface in the grid cell. The value $d = 1$ is selected to approximate a surface robot with a camera a meter above the ground. The highest point in each cell is chosen because visibility generally increases with elevation. This produces a representative point that likely accounts for the greatest exposure case. Finally, we assume that any errors from the discretization can be made negligible by increasing the resolution of the grid. Both environments were tested with $50\!\times\!50$ and $100\!\times\!100$ resolutions.

Each of the presented A*-based algorithms (Exposure Score, Binary, Saturation) and an exposure-agnostic shortest-path A* algorithm, to provide a computational baseline, were tested on both maps with randomized starting and goal locations. Additionally, a high-dimensional best-first-search was run to provide the ground-truth exposure score, all presented paths were able to find a breath-first-search solution. For each combination of starting and goal regions, the A* Saturation algorithm was run with saturation values of $\{1, 2, 3, 4, 5, 10, 15, 20, 25, 50, 100, 200\}$.

Each algorithm's optimality gap for every start-goal region combination is measured as the extra percentage of the map exposed by the potentially sub-optimal path $P_{alg}$ over the percentage exposed by the optimal best-first-search path $P_{bfs}$.
\begin{equation}
    G(P_{alg}, P_{bfs}) = 100 \times \frac{| \epsilon_{alg} | - | \epsilon_{bfs} | }{n}
\end{equation}
where $n$ is the number of regions in the environment and where $\epsilon_{alg}$ and $\epsilon_{bfs}$ are are the set of all regions exposed by the paths $P_{alg}$ and $P_{bfs}$ respectively
\begin{equation}
    \epsilon_{path} = \bigcup_{R_p \in path} \mathcal{E}(p).
\end{equation}

The hardware used to determine the paths for each algorithm consisted of multiple computers with at least 32GB of RAM and the following processors: Intel i7-5930K, Intel i7-12700K, Intel Xeon E5-2670 v3. Custom C\# scripts in the Unity graphics engine \cite{Unity} was used to pre-process the 3D environments. The algorithmic implementations are all single-threaded, written in C++, and compiled with g++ \cite{g_plus_plus_compiler}.

\section{Results}\label{sec:Results}
\begin{figure}[t]
    \vspace{5pt}
    \centering
    \includegraphics[width=.9\linewidth]{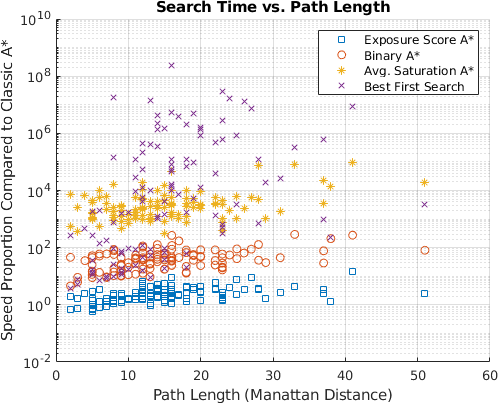}
    \caption{Ratio of computation times for the A* Exposure Search, A* Binary, A* Saturation, and Best-First-Search implementations compared to an exposure-agnostic shortest-path A* implementation. The A* Saturation time is the average of the tested $\tau$ values. \vspace{-1em}}
    \label{fig:distance_vs_search_time}
\end{figure}

The time each algorithm took to compute, normalized by the time for an A* exposure-agnostic search between the locations, is shown in \figref{fig:distance_vs_search_time}. Of note, the A* Exposure Score algorithm is computationally similar to the exposure-agnostic baseline, while the implementations with augmented search nodes, i.e.\,A* Binary and A* Saturation are an order of magnitude slower. The computation time for A* Saturation presented is the average of the computation time for each saturation value tested and is several orders of magnitude slower than the exposure-agnostic computational baseline, but still significantly faster than the best-first-search exposure baseline. For each algorithm, the computation time increases non-linearly as a function of path length, as expected.

\begin{figure}[t]
    \vspace{5pt}
    \centering
    \includegraphics[width=.79\linewidth]{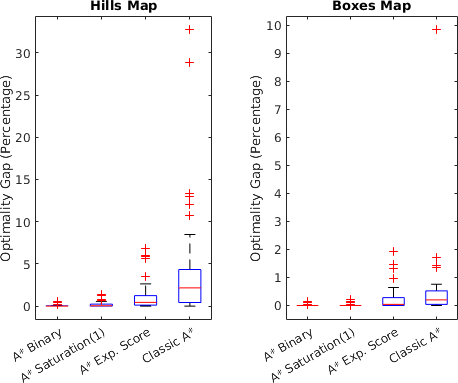}
    \caption{Distribution of optimality gaps of the new A* Exposure Score, Binary, A* Saturation (with $\tau = 1$), algorithms and the Classic shortest-path A* algorithm compared to the BFS ground-truth path exposure for the Hills and Boxes environments. \vspace{-1em}}
    \label{fig:optimality_gaps}
\end{figure}

The distribution of optimality gaps, the difference between the path's exposure and the best-first-search exposure, for the Hills and Boxes maps can be seen in \figref{fig:optimality_gaps}. In both environments, we find that the A* Exposure Score, which performs pre-computation simplifications, requires the least computational time but exhibits the largest optimality gaps of the introduced algorithms. However, it remains reliably better than a standard shortest-path approach. The algorithms with the least optimality gap are A* Binary and A* Saturation when the saturation value $\tau = 1$.

\begin{figure}[b]
    \vspace{-1em}
    \centering
    \includegraphics[width=.9\linewidth]{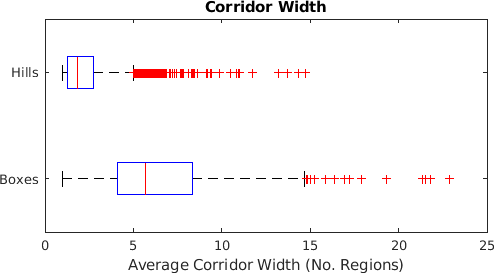}
    \caption{The distributions of average corridor width for all experiment paths in the Hills $(n=2284)$ and Boxes $(n=1386)$ environments. The Boxes map experienced 4 outliers at 41.75, 41.56, 40.50, and 81.63 which correspond to relatively short paths in otherwise open spaces that naturally expose large regions of the map.}
    \label{fig:corridor_width_boxes_and_Hills} 
\end{figure}

The average widths of the corridor computed for every tested path can be viewed in \figref{fig:corridor_width_boxes_and_Hills}. The average corridor width for the Boxes map is several times larger than Hills with median corridor widths of 5.69 and 1.68, respectively, due to the flatness of the Boxes map. Example corridors constructed from several paths can be viewed in \figref{fig:corridor_example_1}.

\section{Discussion}\label{sec:Discussion}
In the environments tested, the performance of the A* Binary search yielded paths with relatively small optimality gaps in exposure, while being computationally similar to exposure-agnostic searches. This result is encouraging, as it affords tractable pre-computation of nearly optimal paths for navigation in the environment. Moreover, the aggregated A* Exposure Score search performed well and efficiently. It could be integrated into on-board planning for systems that need to be exposure-aware while managing dynamic and unstructured scenes. In this implementation, a pre-computed A* Binary solver could generate a path corridor, and then the robot could be deployed with the pre-computed information necessary for A* Exposure Score to do rapid exposure-aware re-planning to adapt to the realities of the environment.

The concept of locally connected equal exposure sets that give rise to the highly nonlinear binary metric of exposure has many uses. These encode all locations a single or team of robots could occupy during a motion without exposing additional regions in the environment. From a robot teaming perspective, this implies that each robot need not follow the same trajectory to minimize exposure, rather they can follow in a corridor and only go single-file in choke points. Further, this affords placing or staging team members at various supporting locations along the path that are minimally exposed.

For example, in robotic avalanche terrain navigation, the goal is quite similar--minimizing exposure to avalanche-prone terrain--and the equal-exposure-corridor sets define where teammates could be stationed to provide oversight of other teammates that are crossing riskier sections of the path.

\begin{figure}[t]
    \vspace{5pt}
    \centering
    \begin{subfigure}[b]{.21\textwidth}
        \centering
        \includegraphics[width=\textwidth]{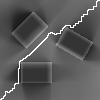}
        \caption{Boxes Path}
        \label{fig:boxes_overview_path}
    \end{subfigure}%
    { }
    \vspace{1em}
    \begin{subfigure}[b]{.21\textwidth}
        \centering
        \includegraphics[width=\textwidth]{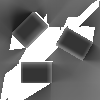}
        \caption{Boxes Corridor}
        \label{fig:boxes_overview_corridor}
    \end{subfigure}
    \centering
    \begin{subfigure}[b]{.21\textwidth}
        \centering
        \includegraphics[width=\textwidth]{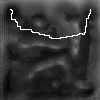}
        \caption{Hills Path}
        \label{fig:hill_overview_path}
    \end{subfigure}%
    { }
    \begin{subfigure}[b]{.21\textwidth}
        \centering
        \includegraphics[width=\textwidth]{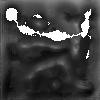}
        \caption{Hills Corridor}
        \label{fig:hill_overview_corridor}
    \end{subfigure}
    \caption{Example seed paths and corridors for the Hills (bottom) and Boxes (top) environments. The corridors show all of the positions or paths that would have no negative contribution to the global exposure metric. The gray-scale value at each pixel maps to the respective region's exposure score \eqref{eqn:exposure_score}, providing an intuitive measure of exposure. \vspace{-1em}}
    \label{fig:corridor_example_1}
\end{figure}

These exposure-corridors can be complemented by the exposure cost gradient. Where the exposure cost gradient is steep, e.g. when taking a corner around an obstacle, the corridor narrows -- more strictly defining the permissible movements. When the exposure gradient is less sensitive to local movements, the corridors tend to widen as can be seen in \figref{fig:corridor_example_1}. Potential movement policies could include fanning out when the corridor is wide to maximize information gain or reduce the risk of team-member collisions, or aligning the individual robots in a straight line when the corridor narrows and the exposure metric requires it. Investigation into the meaning of these sets, on-line path planning, and teaming given these sets is left as future work, but is very motivating for many applications.

The motivating example shown in \figref{fig:motivating_example} illustrates the need to avoid detection, and the presented study is a great first step in this direction. However, the current results are general to geometric line-of-sight exposure to the environment, which is related to but not synonymous with the concept of detection. Detection requires the agent navigating to be both exposed to and noticed by another agent. It also depends on things such as foliage, color contrasts, distance, and the location of the other agent. In future work it would be interesting to explore how an underlying observer location probability distribution and decaying probability of detection with distance would affect the paths found as compared to this geometry-only approach.

\section{Conclusions}\label{sec:Conclusions}

This study highlights the challenges posed by non-Markovian exposure scenarios, such as line-of-sight, in robot path planning. We explored the application of the traditional A* algorithm to navigate in these complex environments focusing on the optimality gaps different Markovian-heuristic approximations experience and introduced the concept of \textit{equal-exposure corridors}. The quantification of the optimality gap provides insights into the trade-offs between algorithm efficiency suitable for embedded systems and the level of optimality achieved. The equal-exposure corridors, like movement-corridors, offer the flexibility for local movement planners to avoid dynamic obstacles by adjusting the robot's trajectory without negatively impacting exposure metrics.

\bibliographystyle{unsrt}
\bibliography{references}

\end{document}